\documentclass{ieeeconf}

\usepackage[pdftex]{graphicx}

\newcommand{\Fig}[1]{Figure~\ref{fig:#1}}

\IEEEoverridecommandlockouts
\begin{document}

\title{Modeling Basic Aspects of Cyber-Physical Systems\thanks{This
    work was supported by the US NSF, Swedish KK-Foundation CERES and
    CAISR Centres, and the Swedish SSF NG-Test Project.}}

\author{
  Walid Taha and Roland Philippsen\\
  Halmstad University, Sweden\\
  Rice University, Houston, TX
}

\maketitle

\thispagestyle{empty}

\begin{abstract}
Designing novel cyber-physical systems entails significant, costly
physical experimentation.  Simulation tools can enable the
virtualization of experiments.  Unfortunately, current tools have
shortcomings that limit their utility for virtual experimentation.
Language research can be especially helpful in addressing many of
these problems.  As a first step in this direction, we consider the
question of determining what language features are needed to model
cyber-physical systems.  Using a series of elementary examples of
cyber-physical systems, we reflect on the extent to which a small,
experimental domain-specific formalism called Acumen suffices for this
purpose.
\end{abstract}

\section{Introduction}

Increasing the computational power of everyday products is
revolutionizing the way we live.  Segways can move us from one
location to another without any need for large, cumbersome, or noisy
vehicles.  Cars can park themselves, and warn us when we are changing
lanes unsafely.  The future appears more fantastic than science
fictions depicted it just a few years ago.  At the same time, the
increasingly tight coupling between computational and physical
mechanisms, often described as cyber-physical systems (CPS), is
creating a challenge for the traditional product development cycles.
For example, car manufacturers are concerned about the amount of
physical testing necessary to assure the safety of a car with a high
degree of autonomy.

Since the beginning of time, physical testing has been the basis for
justified true belief in the qualities of a new product.  A key
ingredient of physical testing is having a collection of specific
usage scenarios.  But the presence of even simple computational
components can make it make it difficult to identify enough usage
scenarios to exercise more than a minute fraction of the possible
behaviors of the system.  These observations are spurring the
developers of cyber-physical systems to rethink the traditional
methods and processes for developing and testing new products.

\subsection{Virtual Experiments and Language Research}

One way to alleviate the testing problem is to use computer
simulations~\cite{carloni-2006} to perform virtual
experiments~\cite{bruneau-2012}.  Virtual testing can be used to
quickly eliminate obviously bad designs.  It can also help build
confidence that a new design can pass test scenarios developed by an
independent party~\cite{jensen-2011}.  Creating a framework for
conducting virtual experiments requires a concerted, interdisciplinary
community effort to address a wide range of challenges, including:

\begin{enumerate}
\item
  Educating designers in the cyber-physical aspects of
  the products they will develop, both in terms of:
  \begin{itemize}
    \item
      How they these aspects are modeled, and
    \item
      What types of system-level behaviors they generate.
  \end{itemize}
\item
  Developing expressive, efficient, and robust modeling and simulation
  tools to support the innovation process.  It is particularly
  important that the underlying models are:
  \begin{itemize}
    \item
      Easy to understand and analyze at each stage of the design
      process, and
    \item
      Easy to reason about across stages.
  \end{itemize}
\item
  Accumulating extensive libraries of component models that are both
  \begin{itemize}
  \item
    Grounded in physical principles and analytic methods, and
  \item
    Validated experimentally.
  \end{itemize}
\end{enumerate}

All three challenges would benefit from better language-based
technologies for describing and simulating cyber-physical systems.
Engineering methods centered around a notion of executable or
\emph{effectively-computable} models can have profound positive impact
on the pace of advancement of knowledge and engineering practice in
cyber-physical systems.

\subsubsection{The Educational Challenge}

For decades, engineering and science education has focused on
providing specialized training within well-defined disciplines.  As a
result, to design an advanced cyber-physical system such as a robot,
we must engage several experts with advanced degrees from a number of
different disciplines, such as mechanical engineering, electrical
engineering, computer science, and biology.  Not only does this
recurring task make it difficult to assemble a team with necessary
expertise for a project, the team may still lack a common language for
discussing key issues that are fundamental to the design of robotic
systems, but that are treated differently across disciplines.

Addressing the educational challenge will require a concerted effort
to break down artificial boundaries between disciplines.  A key step
towards achieving this goal will be to find a {\em lingua franca} (or
``common language'') to communicate about fundamental issues that
recur in the development of a variety of different cyber-physical
systems.  Part of such a language will be a jargon for communicating
efficiently among experts; part will be an appropriate mathematical
formalism.  Language research can be particularly helpful in
developing tools that are closely aligned with executable subsets
(c.f. \cite{fortress, zhu-2012}) of mathematical notations that are
already used by many engineers and scientists but are not available in
mainstream programming languages and tools.

\subsubsection{The Modeling and Simulation Challenge}

The modeling and simulation challenge can also be approached from a
linguistic point of view.  Precise reasoning about models during each
stage of the process can be improved by applying classical
(programming) language design principles, including defining a formal
semantics.  Reasoning across stages can be facilitated by using two
ideas from language design: 1) increasing the expressivity of a
language to support multiple stages in the design process, and 2)
automatically compiling models from one stage to the next to reduce
manual work and opportunities for mistakes as models are translated
from one stage of the design process to the next.

\subsubsection{The Modeling Library Challenge}

It seems reasonable to expect that the dynamics surrounding the
development and use of modeling libraries can be similar to those for
software libraries.  We can envision CPS design as a community process
where there are library providers and library consumers, and where
different libraries and test suites are used to benchmark and evaluate
various offerings.  In such a setting, the interfaces between
components become important, motivating questions related to advanced
linguistic techniques such as types, static checking, unit testing,
contracts, assertions, assume-guarantee reasoning, and blame
assignment, all of which can be expected to play a key role in
addressing this challenge.

\subsection{A Small, Experimental Language for Hybrid Modeling}

To better understand the core linguistic issues that arise in
addressing these challenges, we are developing a modeling language
called Acumen~\cite{taha-2012,acumen-web}.  A key characteristic of
modeling and simulation languages for cyber-physical systems is
supporting hybrid (continuous/discrete) mathematical
models~\cite{carloni-2006}.  Modelica \cite{Modelica} and SimScape are
widely used examples of such languages.  Hybrid modeling can be
supported using a small number of constructs, namely:
\begin{itemize}
\item Ground values (e.g., \verb|True|, \verb|5|, \verb|1.3|, \verb|"Hello"|)
\item Vectors and matrices (e.g., \verb|[1,2]|, \verb|[[1,2],[3,4]]|)
\item
  Object definition (\verb|class C (x,y,z) ... end|)
\item
  Object instantiation and termination (\verb|create|, \verb|terminate|)
\item
  Variable declarations (including a special variable called
  \verb|_3D| for generating visualizations) (\verb|private ... end|)
\item
  Variable derivatives (\verb|x'|, \verb|x''|, ...)
\item
  Continuous assignment (\verb|[=]|)
\item
  Discrete assignment (\verb|=|)
\item
  Conditional statements
  (\verb|if|, and 
   \verb|switch|)
\item
  Expressions and operators on reals (\verb|+|, \verb|-|, ...)
\end{itemize}
It appears that a language with just these features can be helpful in
addressing the educational challenge.  For example, we used such a
language for a term-long project in an eight week course on
cyber-physical systems \cite{taha-lecture-notes}, and seems to have
been received positively.  It appears that being able to concretely
explain a wide range of concerns in a small language has two key
benefits.  The first relates to CPS education.  Using a small language
can help highlight the connections between different concepts, and
avoid the introduction of artificial distinctions between
manifestations of the same concept in different contexts.  The second
relates to language and tools research.  Showing how such a wide range
of cyber-physical phenomena can be captured in a small language helps
emphasize the expressivity of such a small language, and provide a
basis for arguing against the introduction of additional language
features until a compelling case for the addition of such language
features has been made.

These observations inspired us to step back from the details of that
particular teaching experience to reflect on two questions relating to
the challenges described above.  First, what can we view as the basic
cyber-physical aspects of a robot system?  Second, how well can they
be explained in terms of a small hybrid-systems modeling formalism?

\subsection{Contributions}

This paper considers several aspects of cyber-physical systems that
can be seen as common features of cyber-physical systems, and uses
them to reflect on how they can be expressed and illustrated using the
small hybrid systems modeling language introduced above.

Visual and geometric presentation is a critical aspect of analytical
modeling that can hide in plain sight (Section \ref{sec-vis}).
Technically, it is not part of analytical modeling, but it is
indispensable for efficiently understanding both the specification and
the results of a virtual experiment.  From a pedagogic point of view,
the trigonometric reasoning involved in creating visualizations
provides a natural path into geometry of motion (kinematics).  Basic
mechanics and dynamics come next (Section \ref{sec-mech}), and a range
of analytical principles used to model physical systems.  They also
motivate the use of differential equations, which in turn provide much
of the background needed to motivate the discussion of control
(Section \ref{sec-control}).  To help experimentally evaluate the
effect of control, it is useful to consider mechanisms for modeling
disturbances (Section \ref{sec-disturb}).  By introducing these
aspects, we are able to present the simplest possible example of how
to model and test a cyber-physical system.  This allows us to return
back to the physical component and refined it.  A natural way to do
this in the robotics domain is to use ideas from rigid body dynamics
(Section \ref{sec-rigid}).  Similarly, we can refine the model of the
control system by capturing the way in which implementation on a
digital computer introduces both discretization and quantization
effects (Section \ref{sec-discretization}).

After the discussion of the individual aspects has been considered, we
summarize our observations about the language (Section
\ref{sec-discussion}) and conclude.

\section{Visual and Geometric Presentation}\label{sec-vis}

Visual presentation plays an essential role in the design of
cyber-physical systems.  For many people, it is hard to imagine a
robot (and possibly any other design, for that matter) without
conjuring an image of a general physical form.  If we want to replace
physical prototyping with virtual prototyping, visualization becomes a
necessity.  From the educational point of view, this can be
serendipitous, because it can provide an opportunity to introduce
trigonometry, which itself is needed to model geometric features of
objects as well as to work with both the kinematics and dynamics of
physical objects.

\subsection{Drawing 3D Objects}

A small language for hybrid modeling and simulation can be easily
extended with a lightweight mechanism for three dimensionals (3D)
visualization \cite{Yingfu-Thesis}.  In Acumen, the user can specify
3D visualizations through a special variable called \_3D.  This
variable is only special in that it is read by the implementation and
used to generate a dyanmic 3D scene.  In principle, any graphical
rendering technology can be used by an implementation to realize these
visualizations.  In practice, the current implementation used the
Java3D library, which is built on topc of OpenGL.

\begin{figure}
  \centering
  \includegraphics[width=0.5\columnwidth]{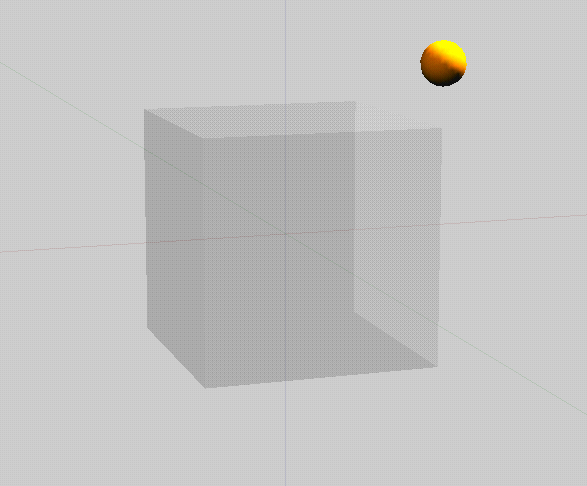}
  \caption{
    The 3D output generated for an instance of the class sphere.
  }\label{fig:sphere-scr}
\end{figure}

\subsection{Class Definitions and Parameterization}

The following class definition specifies a particular way for drawing
a sphere:
\begin{verbatim}
class sphere (m,D)
 private 
  p =[0,0,1];
  _3D = [["Sphere", D+[0,0,1],
          0.03*sqrt(m),
          [m/3,2+sin(m),2-m/2],
          [1,1,1]]];
 end
 _3D [=] [["Sphere", D+p,
           0.03*sqrt(m),
           [m/3,2+sin(m),2-m/2],
           [1,1,1]]];
end

\end{verbatim}
The class parameter m represents a mass.  This parameter is only used
to pick a size and a color for the sphere.  The parameter D is a
display reference point.  Passing different D values to individual
objects facilitates creating visualizations where the individual
objects appear in different places.  The private section declares
local variables as well as their initial value at the (simulated) time
when the object is created.  The variable p is used to represent the
position of the sphere.  The private section and the main body of the
class definition contain similar expressions for \_3D.  Both
expresions consist of a vector that has a format understood by the 3D
visualization part of the Acumen implementation.  The first definition
is a discrete assignment that happens only at object creation time.
The second expression is a continuous assignment that is computed all
the time as long as the object exists in the
simulation.\footnote{Initialization is cumbersome in the current
  syntax for Acumen, as it requires using two very similar
  expressions.}  The format of the vector is as follows: The first
field is a string indicating that the shape we want is a sphere.  The
second field is the coordinate for the center of the sphere.  The next
field is the radius.  Here we compute chose to make the radius a
simple function of the mass.  This function is not intended to have
any physical meaning.  Rather, to produce reasonable effects for the
examples presented in this paper.  The next field contains a vector
that represents the red/green/blue (RGB) colors for this sphere.  To
help us distinguish different objects, we have again used an ad hoc
formula to generate a color based on the mass passed in.  The last
field can be used to express an orientation, and only matters when the
sphere has a texture.  \Fig{sphere-scr} depicts a visualization
generated using this class.

\subsection{Object Creation, Continuous Assignment, and Animation}

We can create sphere by writing ``s = create sphere (5,[0,0,0])'' in
the initialization section and then ``s.p [=] [0.1, 0.2, 0.3]''.  To
generate 3D animations, all we have to do is to let the value of ``p''
vary over time, as in the following code:
\begin{verbatim}
class moving_sphere (m,D)
 private s = create sphere (m,D);
         t = 0; t' = 0
 end
 t'  [=] 5;
 s.p [=] [sin(t)*sqrt(1-(sin(t/10)^2)),
          cos(t)*sqrt(1-(sin(t/10)^2)),
          sin(t/10)];
end
\end{verbatim}
Here the variable t and its derivative t' are introduced here to model
a local variable that progresses at exactly five times the rate of
time.  All that is needed to do that is to include the equation ``t'
[=] 5''.  The time-varying variable t is then used to generate some
interesting values for the x, y, and z components of the the position
field p that represents the center of the sphere object s.

As noted earlier, we can have instances of the same object (such as
the entire moving\_sphere example) appear at different parts on the
screen by varying the D parameter.  By changing the value of the
position parameter p, we can create an animation with two spheres
moving in a synchronized fashion.

It is useful to note that a 3D visualization facility can also be used
to visualize not only 3D values but also scalar values.  For example,
it is useful to define objects that assist in visualizing specific
scalar values during a simulation.  The following class defines a
class to visualize a scalar value as a cylindar of length proportional
to that value:

\begin{verbatim}
class display_bar (v,c,D)
 private 
  _3D = ["Cylinder", D+[0,0.2,0], 
         [0.02,v], c, 
         [-3.14159265359/2,0,0]]
 end
 _3D = ["Cylinder", D+[0,0.2,v/2],
        [0.02,v],c,
        [-3.14159265359/2,0,0]];
end
\end{verbatim}
Following the string ``Cylinder'', the first value represents the
center of the cylinder.  We take this to be v/2 because this will
allow us to keep one end of the cylinder fixed as the value of v
changes.  The next paramter is a tuple containing the radius and
length of the cylinder.  The next parameter is color.  The last
parameter specifies orientation angles for the cylinder.  A screenshot
of an instance of this class will be presented shortly.
\subsection{Vector and Trigonometric Calculation}

In many cases, it is necessary to perform a bit of geometrical
calculation to create the desired shape.  The need for such
calculations can arise in situations that may be simpler than
expected.  An example of such a situation is drawing a cylinder
between two points.  Often, this cannot be done directly because many
underlying visualization tools do not describe cylinders directly in
this exact manner.  Rather, it is common to use two angles that
specify the orientation of the cylinder.  Once we have figured out all
necessary calculations, they can be encapsulated in one class as
follows:

\begin{verbatim}
class cylinder (D)
 private 
  p =[0,0,0]; q=[0,0,0];
  _3D = [["Cylinder", D, [0,0], 
          [0,0,0],[0,0,0]]];
  radius = 0.01; length = 0.01; alpha=0;
  theta= 3.14159265359/2;
  x=0;y=0;z=0
 end
 x [=] dot(p-q,[1,0,0]);
 y [=] dot(p-q,[0,1,0]);
 z [=] dot(p-q,[0,0,1]);
 length [=] norm(p-q);
 alpha [=] asin(z/length);
 if (y>0)
  theta [=] asin(x/(length*cos(alpha)))
 else
  theta [=] -asin(x/(length*cos(alpha)))
             +3.14159265359
 end
 _3D [=] [["Cylinder",(p+q)/2+D,
           [radius,length],
           [1,1,1],[alpha,0,-theta]]];
end
\end{verbatim}
The operators dot and norm operators compute the dot product and the
vector norm (or length).  Creating such an object is a good first
exercise in coordinate transformation.  Versatility with such
transformations is an important skill for working with physical
systems both in terms of Newtonian modeling as well as other, more
advanced modeling techniques (c.f.~\cite{zhu-2012}).  Because they are
executable models that produce visual results, developing small,
purely graphical objects such as the ones above can be a gratifying
way for students to learn about and practice the necessary geometric
and necessary steps to understand how other aspects of robot mechanics
and motion are modeled.

\section{Mechanics and Dynamics}\label{sec-mech}

In contrast to the effort needed to describe geometric and visual
objects, describing basic mechanical systems and their dynamics can be
done more concisely.  A point mass that can only move long dimension
can be represented as follows:
\begin{verbatim}
class mass_1d (m,p0,D)
 private 
  p=p0; p'=0; p''=0; f=0; e_k=0;
  s=create sphere (m,D)
 end
 p'' [=] f/m;
 e_k [=] 0.5 * m * (p')^2;
 s.p [=] [0,0,p]
end
\end{verbatim}
The object takes as parameters a mass m, an initial position p0, and a
reference point for visualization.\footnote{For reasons of space, this
  paper uses short (often single-character) variable names.  While
  this is closer to mainstream mathematical notation, in larger models
  it may be better style to use longer names for variables.}
Internally, the mass keeps track of a position p, its first and second
derivatives p' and p'', a force f, and the kinetic energy e\_k.  For
visualization, a sphere object is created during initialization.  The
body of the class definition specifies that the acceleration of the
object, p'', is determined by Newton's law $F=ma$, where we are
solving for acceleration (which is just p'' here).  The expression for
energy uses the built-in dot-product operation on vectors.  Finally,
we set the position p of the visual object sphere to be the same as
the position p of the current object.

Supporting vector operations make it possible to define a similar
object that has a three dimensional position almost just as simply:
\begin{verbatim}
class mass (m,p0,D)
 private 
  p=p0; p'=[0,0,0]; p''=[0,0,0]; 
  f=[0,0,0]; e_k=0;
  s = create sphere (m,D);
 end
 p'' [=] f/m;
 e_k [=] 0.5 * m * (dot(p',p')) ^2;
 s.p [=] p;
end
\end{verbatim}
Note that it is convenient in this domain to have derivatives over
vectors.  We can induce continuous behaviors in such an object by mean
of an external continuous assignment.  For example, the effect of a
gravitational force on a mass object m by a continuous assignment
``m.f~[=]~m.m*[0,0,-9.81]''.

An idealized, 3D spring can be modeled as follows:
\begin{verbatim}
class spring (k,l0,D)
 private p1=[0,0,0]; p2=[0,0,0];
         f1=[0,0,0]; f2=[0,0,0];
         dl = [0,0,0]; e_p=0;
 end
 dl  [=] p2-p1 * (1-l0/norm(p2-p1));
 f1  [=]  k*dl;
 f2  [=] -k*dl;
 e_p [=] 0.5 * k * dot(dl,dl);
end
\end{verbatim}
This class associates a different force with each end of the spring,
and that computes only a potential energy e\_p rather than a kinetic
energy.  No visualization is included in this object, but that can be
easily done using techniques presented above.

\subsection{Impacts and Discrete Assignment}
\begin{figure}
  \centering
  \includegraphics[width=\columnwidth]{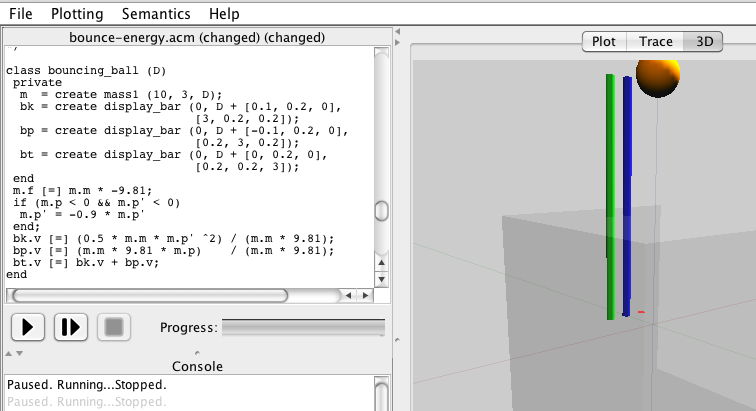}\\[1.1\baselineskip]
  \includegraphics[width=0.3\columnwidth]{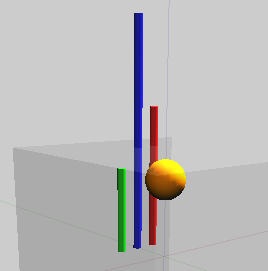}\hfill
  \includegraphics[width=0.3\columnwidth]{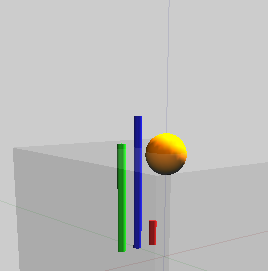}\hfill
  \includegraphics[width=0.3\columnwidth]{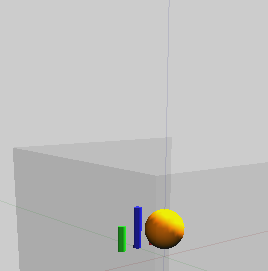}
  \caption{The Acumen IDE with the bouncing ball model
    and simulation results.  The green bar indicates the potential
    energy, the red one is the kinetic energy, and the blue bar is
    their sum.  The total energy decreases with each ground impact,
    and during the free flight phase the two energies behave as
    expected.  }\label{fig:bball-scr}
\end{figure}

An important physical effect in dynamics is impact.  Often, it is
convenient to model impacts as a sudden effect.  Discrete assignments
can be used for this purpose.  The following model provides an example
of the use of discrete assignment to model the impact of a falling
ball with a floor:

\begin{verbatim}
class bouncing_ball (D)
 private
  m  = create mass_1d (10, 3, D);
  bk = create display_bar 
        (0,[3,0.2,0.2],D+[ 0.1,0.2,0]);
  bp = create display_bar 
        (0,[0.2,3,0.2],D+[-0.1,0.2,0]);
  bt = create display_bar 
        (0,[0.2,0.2,3],D+[   0,0.2,0]);
 end
 m.f [=] m.m * -9.81;
 if (m.p < 0 && m.p' < 0)
  m.p' = -0.9 * m.p'
 end;
 bk.v [=] m.e_k / (m.m * 9.81);
 bp.v [=]   (m.m * 9.81 * m.p) 
          / (m.m * 9.81);
 bt.v [=] bk.v + bp.v;
end
\end{verbatim}
The model uses the mass class along with a continuous gravity model
and a ground-impact model where the ball looses 10\% of its velocity.
The class display\_bar is used to display colored bars to present some
additional information in the 3D output.  The mass model used here has
only one degree of freedom along the Z axis.  We use three display
bars to visually represent the kinetic and potential energy, as well
as their sum.  The discrete assignment occurs inside the if statement
that detects impact with the ground plane.  \Fig{bball-scr} shows a
sequence of screenshots, one including the Integrated Development
Environment (IDE), which results from running this example.  It can be
seen that, as expected, the total energy decreases at each impact,
while the kinetic and potential energies reach their respective maxima
and minima at the height of the bounce and the impact at ground level.

Now we turn to creating systems made from components such as the mass
and spring components that we have just introduced.

\section{Capturing Conceptual Structure: Object Boundaries and Composition}\label{sec-obj}

A benefit of using classes in defining a model is that it helps us
think clearly about the conceptual boundary between the different
components that we are modeling.  Connecting components is a matter of
relating fields in different components though continuous assignments.
For example, the following class models a system consisting of three
masses connected by two springs:

\begin{verbatim}
class example_3 (D)
 private 
  m1 = create mass (15,[0,0, 1],D);
  m2 = create mass (5, [0,0,-1],D);
  m3 = create mass (1, [0,0,-1.5],D);
  s1 = create spring (5,1.75,D);
  s2 = create spring (5,0.5,D);
  b  = create display_bar (-1.5,0,D)
 end
 s1.p1 [=] m1.p; s1.p2 [=] m2.p;
 s2.p1 [=] m2.p; s2.p2 [=] m3.p;
 m1.f  [=] s1.f1;
 m2.f  [=] s1.f2 + s2.f1;
 m3.f  [=] s2.f2;
 b.v   [=] (m1.e_k + m2.e_k + m3.e_k
            + s1.e_p + s2.e_p)*12;
end
\end{verbatim}
The class uses an instance of the class display\_bar to draw a
cylinder to display the kinetric energy in the system.  Even though
this is quite a simple dynamical system, it can be used to consider
and illustrate several simple but nevertheless fundamental aspects of
control.
\section{Control}\label{sec-control}
The goal of control is to bring a certain quantity close to a desired
goal.  In the context of the model presented above, and given a
controller object c, the introduction of such a controller can be
modeled as follows:
\begin{verbatim}
// Goal is spring length at rest
 c.g [=] s1.l+s2.l;

// Value is actual spring length
 c.v [=] m1.p-m3.p;
 
// Add c.f
 m1.f [=] s1.f1 + c.f;
 m2.f [=] s1.f2 + s2.f1;

// Subtract c.f
 m3.f [=] s2.f2 - c.f;
\end{verbatim}
In this model the goal value for the controller is to have the length
of the system be the same as the natural lengths of the two springs.
The quantity that we wish to control is the position of the first mass
minus the position of the third one.  The way we will achieve that is
to take a force value f that is generated by the controller and apply
it to both sides of the system that we have constructed, but in
opposing directions.

Now the question that remains is how the controller c should compute
its output force f given the goal g and measured value v.  This is a
prototypical question in the design of control systems, and that can
be approached in a variety of different ways.  Three of the most basic
types of controllers are 1) proportional feedback, 2)
proportional/differential feedback, and 3)
proportion/integral/differential feedback.  The first type can work
for systems without inertia, or that have intrinsic ways of
dissipating inertial energy.  It can be modeled as follows:
\begin{verbatim}
class force_controller_p (k_p)
 private g=[0,0,0]; v=[0,0,0]; 
         f=[0,0,0] 
 end
 f [=] k_p * (g-v)
end
\end{verbatim}
The force f computed is directly proportional (hence the name) to the
difference between the goal g and current value v of the quantity that
we want to control.  The higher the constant k\_p, the higher the
force that will be applied for the same amount of difference (or
``error'') between the goal value and the current value.

If the system has inertia or does not dissipate the extra energy introduced by the control force, it might oscillate indefinitely as a result of the proportional control.
To deal with this problem, a slightly more sophisticated controller that can also add a force opposing the direction of the motion (or rate of change) of the value being measured.
Such a  proportional/differential (PD) controller can be defined as follows:

\begin{verbatim}
class force_controller_pd (k_p,k_d)
 private
  g=[0,0,0]; v=[0,0,0]; s=[0,0,0]; 
  f=[0,0,0]
 end
 f [=] k_p * (g-v) - k_d*s
end
\end{verbatim}
Note that this controller has an extra field s that should be provided
from outside the object to serve as the speed reading that should
affect the final force f.

An interesting feature of these two controllers is that they do not
keep track of history.  We may wish to build a controller that exerts
a higher force only after a weaker force has been tested for some
time.  This can be helpful, for example, if there are external
constant forces (such as gravity) acting on our system, and we do not
know their precise quantity ahead of time.  This type of behavior can
be achieved by a proportional/integral/differential (PID) controller
such as the following:

\begin{verbatim}
class force_controller_pid (k_p,k_i,k_d)
 private
  g=[0,0,0]; v=[0,0,0]; s=[0,0,0]; 
  f=[0,0,0]; i=[0,0,0]; i'=[0,0,0]
 end
 f  [=] k_p*(g-v) + k_i*i - k_d*s;
 i' [=] (g-v)
end
\end{verbatim}
The variable i is being used to integrate the difference between the
goal g and the value v over time, so, no extra inputs are needed.

Using the formalism presented so far, it is easy to simulate and
visualize the several instances of the 3-mass/2-spring example showing
both the behavior of the mass and the energy of the system with
different controllers.  The experiment shows that a P controller will
not dissipate any energy and therefore will not stabilize the system,
and that in fact at times it will add energy to the system and at
others absorb energy from it.  In fact, this example motivates
formally analyzing this system to show that this controller will
function essential as simply another spring between the two extreme
masses.  The PD controller will suffice in stabilizing the system
quickly, and this will be clear from the height of the bar
representing the energy in the system.

\section{Disturbances}\label{sec-disturb}

To enhance the value of an experiment, whether it is a physical or a
virtual experiment, it can be useful to introduce various sources of
disturbance into the system.  At least for preliminary
experimentation, it can be sufficient to model such disturbances as
autonomous sources of various forces.  A simple example that can be
used with the examples presented in this paper is as follows:

\begin{verbatim}
class force_disturbance (k)
 private t=0; t'=0; t''=0; f=[0,0,0] end
 t' [=] 4; f [=] k*[sin(t), cos(t), 
                    sin(2*t+cos(3*t))]
end
\end{verbatim}

This example generates a circular motion in two dimensions, and a
mildly eratic oscillation in the third dimension.  The k parameter is
used to determine the amplitude of the behavior.  One can imagine
further parameterizing the object with a frequency or with
time-varying mixing of the signals along the different axes.  To
determine what type of disturbance is most useful for a particular
class of problems requires experimental analysis and validation of the
models.  For the purposes of this example, the above model sufficies
to give us some confidence that our controllers did not just work for
the particular parameters that we used to test the various systems on.

\section{Rigid Body Dynamics}\label{sec-rigid}

\begin{figure}
  \centering
  \includegraphics[width=0.5\columnwidth]{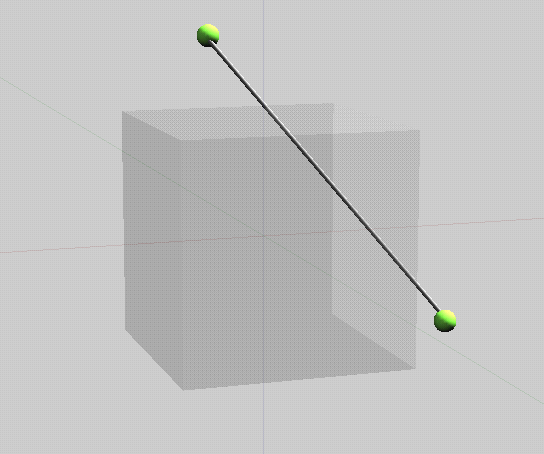}
  \caption{
    The 3D visualization generated by the rod class.
  }\label{fig:rod-scr}
\end{figure}

With basic particle dynamics and control concepts under our belt, we
are ready to start looking at the rigid body dynamics.  A key feature
of this level of analysis of mechanical systems is that we start to
take into account both translational and rotational effects.  It is
also a useful level to illustrate some of the benefits of using vector
algebra to model and reason about the dynamics of systems.  For
example, vector algebra can often allow us to think about problems in
2D and then have the results generalize naturally to 3D.  For example,
consider a rod which holds apart two masses (of m$/2$ each) at a given
distance (visualized in~\Fig{rod-scr}).  Now imagine that there are
force vectors p and q acting on each end.  What is the resulting
acceleration on the system?  The following class models the dynamics
of such a rod:
\begin{verbatim}
class rod (m,p0,q0,D)
 private
  length = norm(p0-q0); 
  p = p0; sp=[0,0,0]; q = q0; sq=[0,0,0];
  axis = (p0-q0)/norm(p0-q0);
  axis'=[0,0,0]; axis''=[0,0,0];
  core = (p0+q0)/2;
  core' = [0,0,0]; core'' = [0,0,0];
  fp = [0,0,0]; fq = [0,0,0]; 
  fp_axis = [0,0,0]; fp_orth =[0,0,0];
  fq_axis = [0,0,0]; fq_orth =[0,0,0];
  c = create dumbbell (1,1,D);
 end
 fp_axis [=]   dot(fp,axis)*axis
             / norm(axis);
 fp_orth [=] fp - fp_axis;
 fq_axis [=]   dot(fq,axis)*axis
             / norm(axis);
 fq_orth [=] fq - fq_axis;
 core'' [=] (fp + fq)/m;
 axis'' [=]   2*(fp_orth-fq_orth)
            / (m*length);
 p [=]   core 
       + (axis * (length/2)/norm(axis));
 q [=]   core 
       - (axis * (length/2)/norm(axis));
 sp [=] core' + axis' * (length/2);
 sq [=] core' - axis' * (length/2);
 c.p [=] p; c.q [=] q;
end
\end{verbatim}
It is easy to combin this system with a controller that works to move
the point p to a predetermined location.  An interesting question is
what happens to the point q during the process, and whether that can
also be controlled as well.  This problem provides a natural starting
point to study challenging questions such as the control of an
inverted pendulum in 3D.

\section{Discretization and Quantization}\label{sec-discretization}
The one aspect of controllers that we have not captured in the models
presented above is that controllers are generally implemented by
digital computers.  The most obvious new issues that result from this
implementation strategy are discretization (in time) and quantization
(in the representation of physical quantities).  Both effects can be
concisely expressed in Acumen.  To model discretization, the key
mechanism that is needed is to define a local clock and to only allow
actions to be performed (or to be observed) at clock transitions.  The
following class models a PID controller (like the one presented above)
with discretization and quantization effects.
\begin{verbatim}
class force_controller_pid_d 
      (k_p,k_i,k_d,period)
 private g=[0,0,0]; v=[0,0,0]; s=[0,0,0]; 
         f=[0,0,0]; t=0; t'=0;
         i=[0,0,0]; i'=[0,0,0] 
 end
 t' [=] 1;
 if (t>period)
  t=0;
  f  [=] k_p*(g-v) + k_i*i - k_d*s;
 end;
 i' [=] (g-v)
end
\end{verbatim}
The variables t and its derivative t' are used in a manner similar to
what was done at the start of this paper to generate an interesting
signal for moving\_sphere.  Here we do two new things with the
variable t.  The first is that we have a conditional statement based
on this variable that waits until (t$>$period).  The parameter period
models the time it takes the particular microprocessor that implements
our controller to produce the new value of the result of the
controller.  Once the condition is true the first thing we do is to
reset the counter.  The second is that we reset its value to 0 using
the statement ``t=0'' as soon as that condition is true.  In addition
to this reset, the conditional also allos the equation for variable f
in the original model to take effect only for that instant when t has
surpassed the value of period.  Because no other definition is given
for this value until this event occurs again (at the start of the next
period), the value f remains constant until that change occurs.

With this model, it is easy to illustrate that as the sampling period
goes up, the system that we are trying to control can become unstable.

Discretization can be modeled by adding another integer (or
fractional) value that is updated (by either a whole integer or a
fraction, depending on the quanta) when an underlying continuous value
goes outside the range represented by the current quanta.

\section{Discussion}\label{sec-discussion}

Developing the examples used in this paper points out several possible
improvements on the current formalism may be necessary.  In this
section, we briefly point out ones that appear to be particular
compelling.

First, when we compose several examples together, the simulation can
slow down.  The formalism is currently implemented as a purely
functional interpreter in Scala.  This serves well the goal of having
a well-defined semantics.  However, we expect that there will be
significant opportunities for improvement in terms of performance.

Second, we have also noticed that numerical stability can be a
concern, even for the simple examples presented here.  The current
implementation uses a forward Euler integrator to simulate the
continuous behavior.  While this is generally considered to be a
simplistic numerical method, it is surprising that it is problematic
even for the simple examples used in this paper.  To address this
problem without becoming dependent on a particular numerical method,
we are investigating the use of interval and enclosure-based methods
to define a semantics for the formalism.

Third, the examples also illustrate that the language can benefit from
improved syntactic support for several different features such as: A)
variable declaration and initialization often seems redundant and/or
verbose, B) quantization for vector-valued variable is currently
verbose, C) discretization could benefit from introducing syntactic
sugar for clocks, D) embedded software could be easier to model if it
can be written directly in a form similar to traditional code.

\section{Conclusions and Future Work}

In this paper we used a small domain-specific modeling formalism aimed
at hybrid systems to express a range of basic aspects of robot
cyber-physics.  By doing so we are able to illustrate how such a
formalism can be a useful basis for learning and communicating about
such concepts.  At the same time, we hope that these examples help
communicate the richness and the power of this formalism despite its
small size.

In future work, we would like to conduct similar studies to determine
whether the same small formalism used here would suffice for
expressing other aspects of robotic cyber-physics, including: more
sophisticated control laws, models for joint and link composition,
collision detection, impact dynamics, a validated model of a
multi-link robot, and a validated model of a team of cooperating
robots.  We anticipate that the improvements suggested in the
discussion section will be important for expressing these aspects
naturally and concisely.

\section*{Acknowledgements}
We would like to thank the reviewers of DSLRob 2012 and Robert
Cartwright for valuable feedback on an earlier draft of this paper.

\bibliographystyle{IEEEtran}
\bibliography{dslrob2012}

\begin{thebibliography}{10}
\providecommand{\url}[1]{#1}
\csname url@samestyle\endcsname
\providecommand{\newblock}{\relax}
\providecommand{\bibinfo}[2]{#2}
\providecommand{\BIBentrySTDinterwordspacing}{\spaceskip=0pt\relax}
\providecommand{\BIBentryALTinterwordstretchfactor}{4}
\providecommand{\BIBentryALTinterwordspacing}{\spaceskip=\fontdimen2\font plus
\BIBentryALTinterwordstretchfactor\fontdimen3\font minus
  \fontdimen4\font\relax}
\providecommand{\BIBforeignlanguage}[2]{{%
\expandafter\ifx\csname l@#1\endcsname\relax
\typeout{** WARNING: IEEEtran.bst: No hyphenation pattern has been}%
\typeout{** loaded for the language `#1'. Using the pattern for}%
\typeout{** the default language instead.}%
\else
\language=\csname l@#1\endcsname
\fi
#2}}
\providecommand{\BIBdecl}{\relax}
\BIBdecl

\bibitem{carloni-2006}
L.~Carloni, R.~Passerone, A.~Pinto, and A.~Sangiovanni-Vincentelli, ``Languages
  and tools for hybrid systems design,'' \emph{Foundations and Trends in Design
  Automation}, vol.~1, no.~1, pp. 1--204, 2006.

\bibitem{bruneau-2012}
J.~Bruneau, C.~Consel, M.~O'Malley, W.~Taha, and W.~M. Hannourah, ``Virtual
  testing for smart buildings,'' in \emph{Proceedings of the 8th International
  Conference on Intelligent Environments (IE'12)}, Guanajuato's, Mexico, 2012.

\bibitem{jensen-2011}
J.~Jensen, D.~Change, and E.~Lee, ``A model-based design methodology for
  cyber-physical systems,'' Istanbul, Turkey, Jul. 2011.

\bibitem{fortress}
E.~Allen, D.~Chase, J.~Hallett, V.~Luchangco, J.-W. Maessen, S.~Ryu, G.~L.
  {Steele Jr.}, and S.~Tobin-Hochstadt., ``The fortress language
  specification,'' Technical report, Sun Microsystems, Inc., 2007.

\bibitem{zhu-2012}
Y.~Zhu, E.~Westbrook, J.~Inoue, A.~Chapoutot, C.~Salama, M.~Peralta, T.~Martin,
  W.~Taha, M.~O'Malley, R.~Cartwright, A.~Ames, and R.~Bhattacharya,
  ``Mathematical equations as executable models of mechanical systems,'' in
  \emph{Proceedings of the First ACM/IEEE International Conference on
  Cyber-Physical Systems}, Stockholm, Sweden, 2012.

\bibitem{taha-2012}
W.~Taha, P.~Brauner, Y.~Zeng, R.~Cartrwright, V.~Gaspes, A.~Ames, and
  A.~Chapoutot, ``A core language for executable models of cyber physical
  systems (preliminary report),'' in \emph{Proceedings of The Second
  International Workshop on Cyber-Physical Networking Systems (CPNS'12)},
  Macau, China, Jun. 2012.

\bibitem{acumen-web}
``Acumen web-site,'' www.acumen-language.org.

\bibitem{Modelica}
P.~Fritzson and P.~Bunus, ``Modelica-a general object-oriented language for
  continuous and discrete-event system modeling and simulation,'' in \emph{SS
  '02: Proceedings of the 35th Annual Simulation Symposium}.\hskip 1em plus
  0.5em minus 0.4em\relax Washington, D.C., USA: IEEE Computer Society, 2002,
  p. 365.

\bibitem{taha-lecture-notes}
W.~Taha, ``Lecture notes on cyber-physical modeling,'' Available online from
  www.effective-modeling.org/p/teaching.html, Sep. 2012.

\bibitem{Yingfu-Thesis}
Y.~Zeng, ``Lightweight three-dimensional visualization for hybrid systems
  simulation,'' Master's thesis, Halmstad University, Halmstad, 2012.

\end{thebibliography}

\end{document}